\documentclass[10pt,twocolumn,letterpaper]{article}

\usepackage{soul}
\usepackage{color}
\usepackage{tabularx}
\usepackage{tabulary}
\usepackage{cvpr}
\usepackage{times}
\usepackage{epsfig}
\usepackage{graphicx}
\usepackage{amsmath}
\usepackage{amssymb}
\usepackage{caption}
\usepackage{float}
\usepackage{multirow}
\usepackage{stackengine}
\usepackage{calc}
\usepackage{stfloats}
\DeclareMathOperator{\atantwo}{atan2}
\usepackage[pagebackref=true,breaklinks=true,letterpaper=true,colorlinks,bookmarks=false]{hyperref}
\usepackage{booktabs}

\newcolumntype{L}[1]{>{\raggedright\let\newline\\\arraybackslash\hspace{0pt}}m{#1}}
\newcolumntype{C}[1]{>{\centering\let\newline\\\arraybackslash\hspace{0pt}}m{#1}}
\newcolumntype{R}[1]{>{\raggedleft\let\newline\\\arraybackslash\hspace{0pt}}m{#1}}

\newcommand{\ignore}[1]{}

\makeatletter
\DeclareRobustCommand\onedot{\futurelet\@let@token\@onedot}
\def\@onedot{\ifx\@let@token.\else.\null\fi\xspace}

\def\eg{e.g\onedot}

\def\etal{et al\onedot}

\makeatother

\definecolor{MyDarkBlue}{rgb}{0,0.08,1}
\definecolor{MyDarkGreen}{rgb}{0.02,0.6,0.02}
\definecolor{MyDarkRed}{rgb}{0.8,0.02,0.02}
\definecolor{MyDarkOrange}{rgb}{0.40,0.2,0.02}
\definecolor{MyPurple}{RGB}{111,0,255}
\definecolor{MyRed}{rgb}{1.0,0.0,0.0}
\definecolor{MyGold}{rgb}{0.75,0.6,0.12}
\definecolor{MyDarkgray}{rgb}{0.66, 0.66, 0.66}

\newcommand{\myparagraph}[1]{\vspace{-11pt}\paragraph{#1}}
\newcommand{\myitem}{\vspace{-5pt}\item}

\cvprfinalcopy %

\def\cvprPaperID{8495} 

\ifcvprfinal\fi
\begin{document}

\title{End-to-End Optimization of Scene Layout}
\author{Andrew Luo\textsuperscript{1} \quad Zhoutong Zhang\textsuperscript{2} \quad Jiajun Wu\textsuperscript{3} \quad Joshua B. Tenenbaum\textsuperscript{2} \\ 
\textsuperscript{1}Carnegie Mellon University\qquad \textsuperscript{2}Massachusetts Institute of Technology\qquad \textsuperscript{3}Stanford University
}


\maketitle
\raggedbottom
\begin{abstract}

We propose an end-to-end variational generative model for scene layout synthesis conditioned on scene graphs. Unlike unconditional scene layout generation, we use scene graphs as an abstract but general representation to guide the synthesis of diverse scene layouts that satisfy relationships included in the scene graph. This gives rise to more flexible control over the synthesis process, allowing various forms of inputs such as scene layouts extracted from sentences or inferred from a single color image. Using our conditional layout synthesizer, we can generate various layouts that share the same structure of the input example. In addition to this conditional generation design, we also integrate a differentiable rendering module that enables layout refinement using only 2D projections of the scene. Given a depth and a semantics map, the differentiable rendering module enables optimizing over the synthesized layout to fit the given input in an analysis-by-synthesis fashion. Experiments suggest that our model achieves higher accuracy and diversity in conditional scene synthesis and allows exemplar-based scene generation from various input forms. 
\end{abstract}
\section{Introduction}

Interior scene layout generation is primarily concerned with the positioning of objects that are commonly encountered indoors, such as furniture and appliances. It is of great interest due to its important role in simulated navigation, home automation, and interior design. The predominant approach is to perform unconditioned layout generation using implicit likelihood models~\cite{yu2011make}. These unconditional models can produce diverse possible layouts, but often lack the fine-grained control that allows a user to specify additional requirements or modify the scene. In contrast to the unconditional models, conditional layout generation uses various types of inputs, such as activity traces, partial layouts, or text-based descriptions, enabling more flexible synthesis. 

In this work, we use the 3D scene graph representation as a high-level abstraction, with the graph not only encoding object attributes and identities, but also 3D spatial relationships. Not only does this design enable more control over the generated content, as users can directly manipulate the input scene graph, it also serves as a general intermediate representation between various modalities of scene descriptions, such as text-based descriptions and exemplar images. 

\begin{figure}[t]
    \centering
    \includegraphics[width=\linewidth]{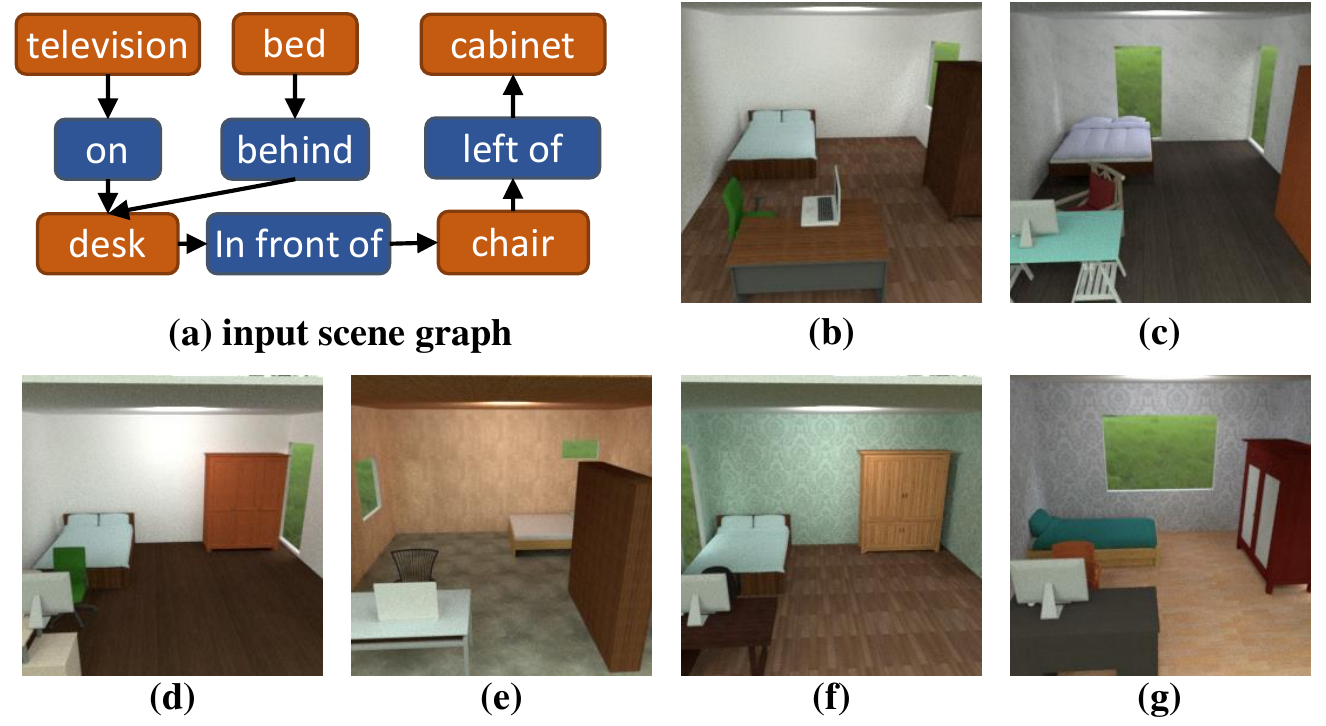}
    \vspace{-20pt}
    \caption{Conditional scene synthesis. (a) The input is a scene graph describing object relationships. (b)--(g) Diverse layouts synthesized conforming to the input scene graph.}
    \vspace{-15pt}
    \label{fig:teaser}
\end{figure}

Many previous methods have used scene graphs as an intermediate representation for downstream vision tasks such as image synthesis, where they mostly formulate the problem in a deterministic manner. In contrast, our model respects the stochastic nature of the actual scene layout conditioned on the abstract description of a scene graph. The model we introduce, named 3D Scene Layout Network (3D-SLN), is a general framework for scene layout synthesis from scene graphs. 3D-SLN combines a variational autoencoder with a graph convolutional network to generate diverse and plausible layouts that are described by the relationships given in the 3D scene graph, as shown in Figure~\ref{fig:teaser}.

We further demonstrate how a differentiable renderer can be used to refine the generated layout using a single 2.5D sketch (depth/surface normal) and the semantic map of a 3D scene. In addition, our framework can be applied to perform exemplar-based layout generation, where we synthesize different scene layouts that share the same scene graph extracted from text or inferred from a reference image.

In summary, our contributions are threefold. First, we introduce 3D-SLN, a conditional variational autoencoder--based network that generates diverse and realistic scene layouts conditioned on a scene graph. Second, we demonstrate our model can be fine-tuned to generate 3D scene layouts that match the given depth and semantic information. Finally, we showed that our model can be useful for several applications, such as exemplar-based layout synthesis and scene graph--based image synthesis. 

\section{Related Work}

Our method is related to the multiple areas in computer vision and graphics, including scene graph representations, scene synthesis, and differentiable rendering.

\myparagraph{Scene graphs.}
Scenes can be represented as scene graphs---directed graphs whose nodes are objects and edges are relationships between objects. Scene graphs have found wide applications such as image retrieval~\cite{johnson2015image} and image captioning~\cite{anderson2016spice}. There have also been attempts to generate scene graphs from text~\cite{schuster2015generating}, images~\cite{xu2017scene, newell2017pixels, li2017scene}, and partially completed graphs~\cite{wang2019planit}. In this paper, we use scene graphs to guide our synthesis of 3D indoor scene layout.

\myparagraph{Scene synthesis.}
In computer graphics, there has been extensive research on indoor scene synthesis. Much of this work is associated with producing plausible layouts constrained by a statistical prior learned from data. Typical techniques used include probabilistic models~\cite{fisher2012example, fisher2015activity, yu2011make}, stochastic grammar~\cite{qi2018human}, and recently, convolutional networks~\cite{wang2018deep, ritchie2019fast}.

Many of these approaches build upon the recent advancement of large-scale scene repositories~\cite{song2017semantic, mccormac2017scenenet, zheng2019structured3d} and indoor scene rendering methods~\cite{zhang2017physically, li2018interiornet,jiang2018configurable}. These methods typically focus on modeling the possible distribution of objects given a particular room type (\eg, bedrooms). Some recent papers~\cite{chang2015text} have studied 3D scene layout generation directly from text. Some concurrent work also uses relational graphs for modelling scenes, however our approach is capable of single-pass scene synthesis in a fully differentiable manner, where as~\cite{wang2019planit} tries to generate the scene graph in an autoregressive fashion, and has an non-differentiable sampling step. 

\myparagraph{Differentiable rendering. } 
Traditional graphics engines do not produce usable gradients for optimization purposes. A variety of renderers that allow for end-to-end differentiation have been proposed~\cite{loper2014opendr, li2018differentiable, kato2018neural, liu2019soft}. These differentiable renderers have been used for texture optimization~\cite{kato2018neural}, face inference~\cite{tewari2018self}, single image mesh reconstruction~\cite{henderson2018learning}, and scene parsing~\cite{huang2018holistic}. Additionally non-differentiable rendering has been used with approximated gradients~\cite{3DRCNN_CVPR18} for instance level 3D construction. We utilize the neural mesh renderer~\cite{kato2018neural}, which allows us to manipulate the layout and rotation of individual objects given depth and semantic maps as reference.

\section{Methods}

\begin{figure*}[t]
  \centering
  \includegraphics[width=0.9\linewidth]{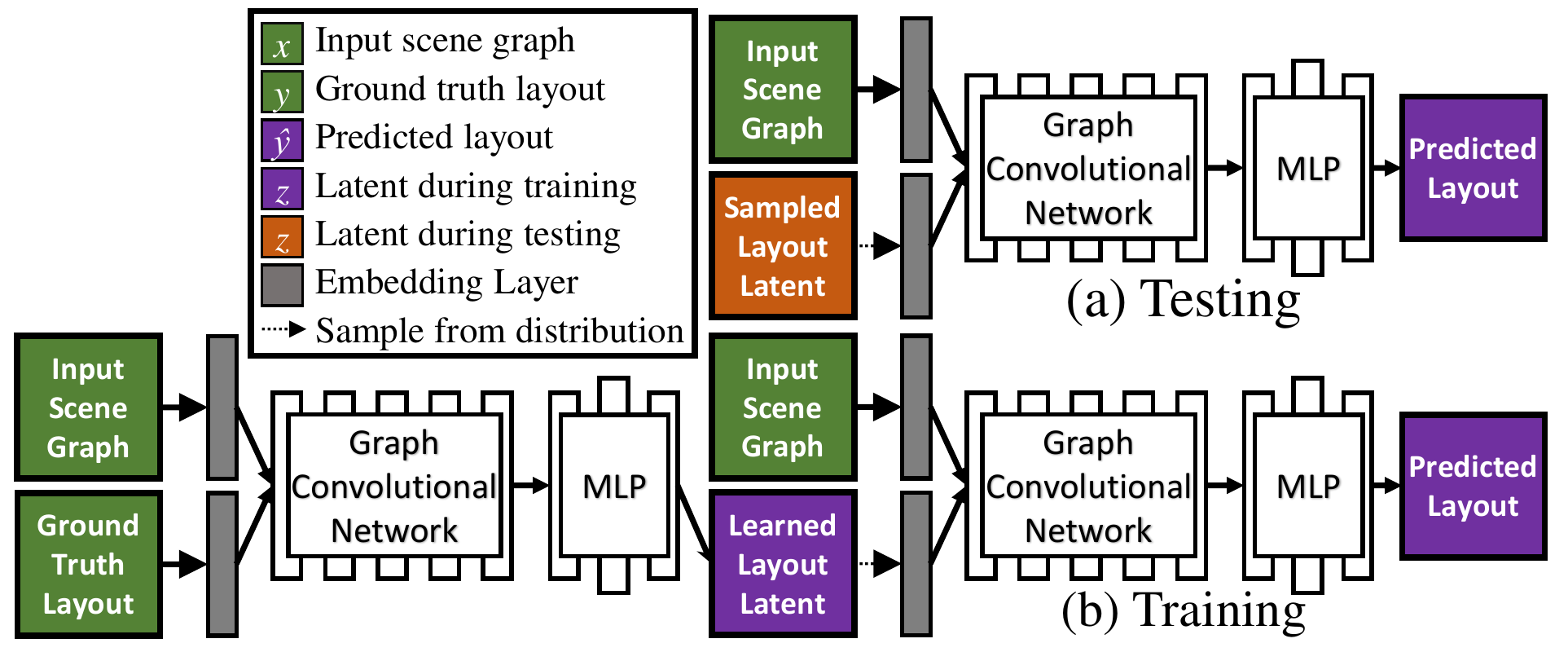}
  \vspace{-0.5em}
  \caption{Network architecture of the scene layout generator. (a) At test time, a latent code is sampled from a learned distribution and is sent to a decoder with the scene graph to generate scene layout. (b) During training, an encoder converts the ground truth scene layout and the scene graph into a distribution, from which the latent code is sampled and decoded.}
  \vspace{-1.5em}
  \label{layoutnet}
\end{figure*}

We propose \textbf{3D Scene Layout Networks} (3D-SLN), a conditional variational autoencoder network tailored to operate on scene graphs. We first use a graph convolution network to model the posterior distribution of the layout conditioned on the given scene graph. Then, we generate diverse scene layouts, which includes each object's size, location and rotation, by sampling from the prior distribution.

\subsection{Scene Layout Generator}
While previous methods generate 2D bounding boxes from a scene graph~\cite{johnson2018image} or text descriptions~\cite{hong2018inferring}, our model generates 3D scene layouts, consisting of the 3D bounding box and rotation along the vertical axis for each object. In addition, we augment traditional 2D scene graphs to \textit{3D scene graphs}, encoding object relationships in 3D space. 

Specifically, we define the $X$ and $Y$ axis to span the plane consisting of the room's floor, and an up-direction $Z$ for objects above the floor. Under such definition, the relationship `left of' constraints the $X$ and $Y$ coordinates between pairs of objects, while the relationship `on' constraints the $Z$ coordinate between them. Each node in the scene graph will not only define what \textit{type} of object it is, it may optionally define object's attributes regarding the object height \textit{(tall, short)} but also volume \textit{(large, small)}. The scene graph $y$ is represented by a set of relationship triplets, where each triplet is in the form of $(o_{i},p,o_{j})$. Here $p$ denotes the spatial relationship and $o_{i}$ denotes the $i$-th object's type and attributes.

In order to operate on the input graph and to generate multiple scenes from the same input, we propose a novel framework, named 3D Scene Layout Network (3D-SLN), combining a graph convolution network (GCN)~\cite{johnson2018image} with a conditional variational autoencoder (cVAE)~\cite{sohn2015learning}. The architecture is shown in Figure~\ref{layoutnet}. During training, the encoder is tasked to generate the posterior distribution of a given scene layout conditioned on the corresponding scene graph. The encoder therefore takes a scene graph and an exemplar layout as input, and outputs the posterior distribution for each object, represented by the mean and log-variance of a diagonal Gaussian distribution. A latent vector is then sampled from the Gaussian for each object. The decoder then takes the sampled latent vectors and the scene graph as input and generates a scene layout, represented by the 3D bounding box and its rotation for each object.

We define $x$ to be the input scene graph, $y$ to be a exemplar layout, $\hat{y}$ to be the generated layout, and $\theta_e, \theta_d$ to be the weights of the encoder $P_{\theta_e}$ and decoder $Q_{\theta_d}$ of 3D-SLN, respectively. Each element in $y_i$ in layout $y$ is defined by a 7-tuple, representing the bounding box and the rotation of each object $i$:
\begin{equation}
\small
   y_i = (\texttt{min}_{X_i}, \texttt{min}_{Y_i},\texttt{min}_{Z_i}, \texttt{max}_{X_i},\texttt{max}_{Y_i},\texttt{max}_{Z_i}, \omega_i),
\end{equation}
where $\texttt{min}_{X_i}, \texttt{min}_{Y_i},\texttt{min}_{Z_i}, \texttt{max}_{X_i},\texttt{max}_{Y_i},\texttt{max}_{Z_i}$ denotes the 3D bounding box coordinates, and $\omega_i$ denotes the rotation around the $Z$ axis.

To train the graph-based conditional variational autoencoder described above, we optimize
\begin{equation}
\small
\mathcal{L}(x,y;\theta) =\lambda D_{KL}(P_{\theta_e}(z|x,y)|p(z|x))+{L_{\text{layout}}}(Q_{\theta_d}(x,z),y),
\end{equation}
where $\lambda$ is the weight of the Kullback-Liebler divergence, $p(z|x)$ is the prior distribution of the latent vectors, which is modeled as diagonal Gaussian distribution, and $\mathcal{L}_{\text{layout}}$ is the loss function defined over layouts. $\mathcal{L}_{\text{layout}}$ consists two parts: $\mathcal{L}_{\text{position}}$ and $\mathcal{L}_{\text{rotation}}$. $\mathcal{L}_{\text{position}}$ is defined as the $L1$ loss over each object's bounding box parameters. For the rotation, we first discretize the range of the rotation angles to 24 bins, and define $\mathcal{L}_{\text{rotation}}$ as the negative log-likelihood loss between the discretized angles for all the objects. We apply learned embedding layers to process object type, rotation, attribute, and relations; and a linear layer to process the bounding box. The rotation and box embeddings are used for the encoder only. The object type, bounding box, rotation, attribute, and relational embeddings have dimensions $[48, 48, 16, 16, 128]$. Embeddings are computed separately for the encoder and decoder. The intermediate latent representation is a 64 dimensional vector for each object. Both the encoder and decoder contain five graph convolution layers with average pooling and batch normalization. 

At test time, we use the decoder to sample scene layouts from scene graphs. We first sample latent vectors from the prior distribution, modeled as a Gaussian distribution. Given the sampled latent vectors and the 3D scene graph, the decoder then generates multiple possible layouts.

\subsection{Gradient-Based Layout Refinement}

Here we consider the case where we would like to generate a layout that fits a target layout, represented as an depth image $D$ with corresponding semantics $S$. Using our scene graph and an inferred layout, we first retrieve object meshes from the SUNCG dataset to construct a complete scene model. Specifically, for each object $i$ in the generated layout, we retrieve its 3D model $M_i$ from the SUNCG dataset by finding the models with the most similar bounding box parameters within its class.

After instantiating a full 3D scene model, we then utilize a differentiable renderer~\cite{kato2018neural} $R$ to render the corresponding semantic image $\widetilde{S}$ and the depth image $\widetilde{D}$ from the scene. The rendered images are then used to compare with the target semantics and depth image. This provides the gradients to update both the sampled latent vectors and the weights of the decoder, making the generated 3D layout to be consistent with the input semantics and depth.

Specifically, we note the entire generate process as
\begin{align}
\widetilde{S} &= R_S(\hat{y}_1,M_1,\hat{y}_2,M_2,...,\hat{y}_N,M_N), \\
\widetilde{D} &= R_D(\hat{y}_1,M_1,\hat{y}_2,M_2,...,\hat{y}_N,M_N), \\
\hat{y} &= Q_{\theta_d}(x,z),
\end{align}
where $R_S$ denotes the rendered semantic map, $R_D$ denotes the rendered depth map, $\hat{y}$ denotes the generated layout, and $N$ denotes the number of objects in the scene graph $x$.

To optimize the decoder and the latent vectors, we aim to calculate the gradient of $\widetilde{D}$ and $\widetilde{S}$ with respect to $z$ and $\theta_d$. Note that since the output of $\omega_i$ in $y_i$ is discretized into 24 bins, i.e. $\omega_i = \text{bin}_k$, where $k = \text{argmax}_k(\{\omega_{ik}|k\in[1,2,...24]\})$, the entire rendering process is not differentiable due to the $\text{argmax}$ operator. To overcome this, we use a softmax-based approximation to compute the angle for the $i$-th object, defined as (assuming a $k$-way prediction)\vspace{-5pt}
\begin{align*}
    x_k &= \sum_{n=1}^{k} 1,\quad \omega_{ik} = \frac{e^{\omega_{ik}}}{\sum_{k}{e^{\omega_{ik}}}},\\
    \omega_i &= \sum{(\omega_{ik} \times x_k)} -1.0.
    \vspace{-5pt}
\end{align*}
By doing so, we can take gradients of the loss function between the rendered images $\widetilde{D},\widetilde{S}$ and the target images $D,S$. A simple loss function can be defined as $\mathcal{L}_{\text{total}} = \mathcal{L}_{2}(D,\widetilde{D})+\mathcal{L}_{\text{cross-entropy}}(S,\widetilde{S})$. This however, leads to highly unstable gradients in practice. To stabilize the layout refinement process, we calculate the loss between the depth images in a per-class manner. More specifically, for each class $c$, we calculate its class-conditioned depth map $D_C$ as \vspace{-5pt}
\begin{align}
    D_c[S==c] &= D \otimes S[S==c],\\
    D_c[S \neq c] &=  \text{mean}(D_{c}[S==c]).
    \vspace{-5pt}
\end{align}
That is, we keep the depth values that lies within a particular semantic class $c$, and fill the rest of the values with the mean depth of this class.
Therefore, we rewrite the depth loss as $L_\text{depth} = \sum_c \mathcal{L}_2(\widetilde{D}_c,D_c)$.

This can be understood as a class-wise isolation of the depth gradient, and can prevent spurious optima during the layout refinement process. We also impose a soft constraint on the change in object sizes with an additional $\mathcal{L}_{1}$ penalty on the size of each object during optimization at a given time step $\widetilde{s}_t$ compared to the original size $\widetilde{s}_{0}$. To facilitate the optimization process when target and proposed layouts may not have an exact match in shape, we apply multi-scale average pooling on both the candidate and target. The total refinement loss with respect to a target depth and semantic is $\mathcal{L}_{\text{total}} = \alpha \mathcal{L}_{\text{size}} + \beta \mathcal{L}_{\text{depth}} + \gamma \mathcal{L}_{sem}$, where $\alpha$, $\beta$, and $\gamma$ are hyper-parameters, and $\mathcal{L}_\text{size} =|\widetilde{s}_t, \widetilde{s}_0|_1$, $\mathcal{L}_\text{depth} = \sum_c |\widetilde{D}_c, D_c|_2^2$, and
$\mathcal{L}_\text{sem} = \mathcal{L}_{\text{cross-entropy}}({\widetilde{S},S})$.

\begin{figure}[t]
  \centering
  \includegraphics[width=\linewidth]{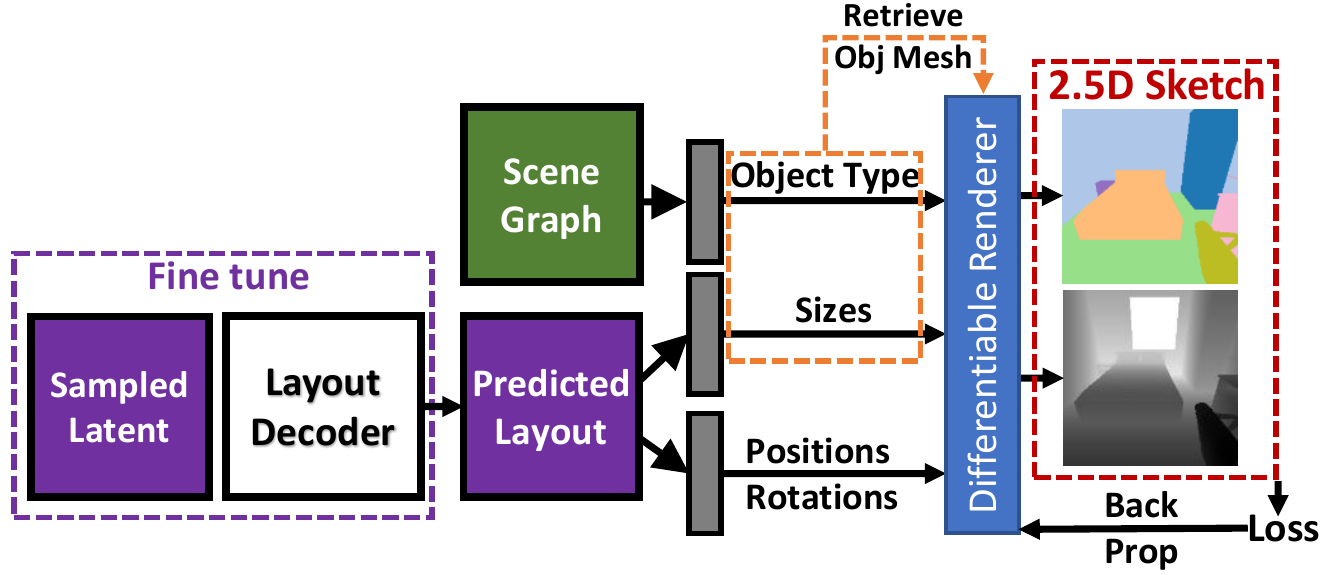}
  \vspace{-1.5em}
  \caption{We can fine-tune object positions, sizes, and rotations by computing the difference in estimated and ground-truth 2.5D sketches \& semantic maps, and back-propagating the gradients.}
  \vspace{-1.5em}
  \label{fig:neuralmesh}
\end{figure}

This allows us to obtain meaningful gradients from a exemplar 2D projection of a scene to optimize $\hat{y}$ by calculating a gradient with respect to the sampled latent vector $z$ and the decoder. The framework for our gradient based layout refinement is shown in Figure~\ref{fig:neuralmesh}.

\section{Experiments}

In this section, we compare our approach with state-of-the-art scene layout synthesis algorithms to demonstrate the quality and diversity of our synthesized scenes. Additional ablation studies show that each component in our model contributes to its performance. Finally, we demonstrate our algorithm also enables exemplar based layout synthesis and refinement.

\subsection{Setup}

For layout generation, we learn from bedroom layouts in SUNCG~\cite{song2017semantic}. The training dataset consists of 53,860 bedroom scenes with 13.15 objects in each scene on average. During training, we use synthetic scene graphs sampled from the ground truth scene layout, which can avoid human labeling and also serve as data augmentation. At test time, we can either use human-created scene graphs or sample scene graphs from the validation set as model input.

The cVAE 3D graph network is trained on a total of $600k$ batches, which takes around $64$ hours with a single \texttt{Titan Xp}. For each batch we sample 128 scene graphs. A learning rate of $10^{-4}$ is used with the Adam optimizer. We use three losses with the following weights: $\lambda_{pos}=1, \lambda_{rot}=1, \text{and }  \lambda_{KL}=0.1$. 
\subsection{Scene Layout Synthesis}

We first evaluate our 3D-SLN on scene layout synthesis from a scene graph. 
We sample 10 layouts per scene, and calculate the average standard deviation for object size, position, and rotation. Layout synthesis alone during testing is highly efficient, taking about $70$ms on a GPU for a batch of 128 graphs.

\begin{table*}[t!]
    \centering
    \begin{tabular}{lccccc}
        \toprule
        Model            & Scene Graph Acc. (\%) & $\mathcal{L}_1$ box loss & STD (size) & STD (position) & STD (rotation) \\ 
        \midrule
        Random Layout    & 57.1             & 0.317       & 0.000       & 0.244          & 6.48        \\
        Perturbed Layout & 82.6             & 0.080       & 0.000       & 0.100          & 3.00        \\ 
        \midrule
        DeepSynth        & N/A              & N/A         & N/A        & 0.129          & 2.27        \\
        \midrule
        GCN              & 86.3              & 0.111         & 0.000        & 0.000            & 0.00         \\
        GCN+noise        & 86.9              & \textbf{0.109}         & 0.001        & 0.002            & 0.18         \\
        \midrule
        3D-SLN (Ours)  & \textbf{94.3}             & 0.148       & \textbf{0.026}      & \textbf{0.078}          & \textbf{4.77} \\
        \bottomrule
    \end{tabular}
    \vspace{-0.5em}
    \caption{Quantitative results on scene layout generation. We use scene graph accuracy and $\mathcal{L}_{1}$ bounding box loss to evaluate the accuracy of generated scene layouts. Standard deviation of boxes and angles are used to measure the diversity of scene layouts. In the above evaluation, bounding boxes are normalized in the range $[0,1]$, while angles are represented as integers ranging from 0 to 23. DeepSynth~\cite{wang2018deep} is used as a baseline.}
    \label{tab:1}
    \vspace{-10pt}
\end{table*}

\myparagraph{Baselines.} 
We compare our model with the state-of-the-art scene layout synthesis algorithm, DeepSynth~\cite{wang2018deep}. Following Qi~\etal\cite{qi2018human}, we also include two additional baselines: Random, where every object is distributed randomly in a room; and Perturbed, where we perturb object positions against their ground truth positions with a variance of 0.1 on their spatial location (all locations are normalized to $[0,1]$) and with a standard deviation of 3 bins on their rotation (approximately 0.785 radians).

\myparagraph{Metrics.}

We analyze both the accuracy and diversity of the results through three metrics:
\begin{itemize}
    \myitem Scene graph accuracy measures the percentage of scene graph relationships a given layout respects, and is a metric that measures input-output alignment. 
    \myitem $\mathcal{L}_1$ loss of the proposed and ground truth bounding boxes. It should be noted that since the goal is the generate multiple plausible layouts, $\mathcal{L}_1$ is not necessarily a meaningful metric and is provided for reference only. 
    \myitem The standard deviation of the size, position, and rotation of objects in predicted scene layouts. Because DeepSynth produces layouts in an autoregressive fashion, a particular object of interest (\eg, a bed) might appear at various steps across multiple trials. Due to the lack of correspondence, we can only compute the standard deviation for all objects within each semantic category, and average the standard deviations across all categories. For our model and the random/perturbed layout models, we calculate standard deviations for each object of interest and compute their mean. 
\end{itemize}

\myparagraph{Results.} 
Table~\ref{tab:1} shows that our model has the highest scene graph accuracy and diversity. This indicates that our model has successfully learned to position objects according to a distribution rather than approximating a fixed location. While DeepSynth has a higher standard deviation in object positions, it has a lower standard deviation in rotations. It also does not allow fine-grained control of the synthesis process. Although `Perturbed Layout' has the lowest $\mathcal{L}_1$ loss, it has a significantly lower scene graph accuracy. 

\subsection{Ablation Study}
We perform ablation studies on our scene layout generation network. By utilizing a graph convolution network combined with a VAE, it is able to generate multiple plausible layouts from a given scene graph. 

\myparagraph{Baselines.}
Following Johnson~\etal\cite{johnson2018image}, we run an ablated version of our network (denoted as GCN) that consists of a single graph convolution network followed by an MLP to predict the layout conditioned on a scene graph. This baseline is deterministic. We also propose a different method, GCN+noise, which samples noises from $\mathcal{N}(0,1)$ to perturb the layout of the GCN baseline.

\myparagraph{Results.} 
Table~\ref{tab:1} shows that our full model (3D-SLN) achieves the highest scene graph accuracy, indicating that most of the synthesized scene layouts indeed follow the input scene graph. Our full model also achieves the highest diversity, as measured in the standard deviation in the size, position, and rotation of objects in the synthesized scene.

\begin{figure*}[t]
  \centering
  \includegraphics[width=0.9\textwidth]{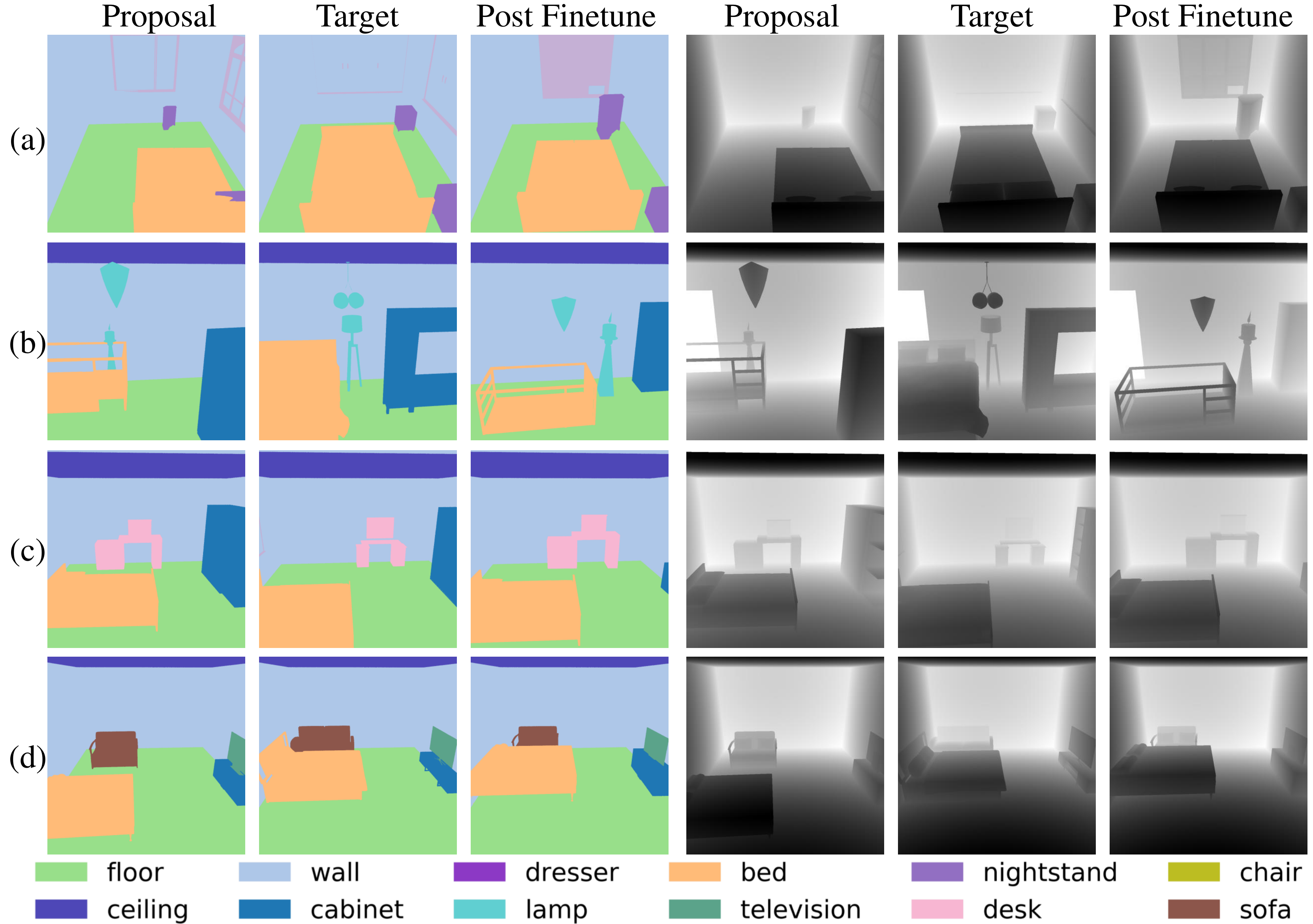}
  \vspace{-0.5em}
  \caption{Each row shows a test case for exemplar-based layout fine-tuning. The left three columns represent the 2D semantic map of the initially proposed layout, the ground truth target, and the semantic map after layout optimization. The right three columns represent the 2.5D depth map similarly. \textbf{(a)} After fine-tuning, the bed has moved to the center as in the target, and  the two night stands become more prominent; \textbf{(b)} The lamp (in light blue) has moved downwards; \textbf{(c)} The desk has moved to the right after optimization and the bed has become closer; \textbf{(d)} The bed has moved closer to the sofa. }
  \vspace{-1em}
  \label{finetunevisual}
\end{figure*}

\begin{table}[t]
    \centering\small
    \begin{tabular}{lccc}
        \toprule
        Metrics & Pre-Finetune & Post-Finetune & Improve (\%) \\
        \midrule
        3D IoU & 0.2353 & 0.3035 & 28.9\\ 
        Depth MSE & 0.0525 & 0.0480 & 8.64 \\ 
        Semantic CE & 2.9471 & 2.8504 & 3.28 \\ 
        \bottomrule
    \end{tabular}
    \vspace{-5pt}
    \caption{Quantitative results on finetuning with 2.5D sketches of a target layout. We measure the Intersection-over-Union (IoU) of the 3D bounding boxes, class-specific mean squared error (MSE) of the depth maps, as well as the cross-entropy loss (CE) on semantic maps.}
    \label{finetune_number}
    \vspace{-10pt}
\end{table}

\subsection{End-to-End Layout Refinement}

We now demonstrate that our model can be guided by 2.5D sketches and semantic maps when synthesizing 3D scene layout from a scene graph. We perform optimization over 150 randomly selected scene graphs. Here, for sampling different scene layouts from our stochastic model, we use the latent sampled from the ground truth bounding boxes of a given scene. To prevent cases when wall or object occlusion negatively impact optimization performance, we take six attempts at a given scene graph, and select the best. The analysis-by-synthesis process requires a forward (rendering) pass to produce depth and semantic maps, then a backwards pass to produce gradients. We optimize  for 60 steps, taking three minutes for each scene on average.

\myparagraph{Metrics.}
We use three metrics for this problem: The first is done in 3D, and captures the Intersection-over-Union (3D IoU) of objects and their target after all transformations (rotations and translations) are applied. The latter two are performed in the 2D projection: we calculate a mean-squared error (MSE) on the predicted and ground-truth depth maps; we also calculate the cross-entropy (CE) loss of our current proposed layout against the target layout. 

\begin{figure}[t]
  \includegraphics[width=\linewidth]{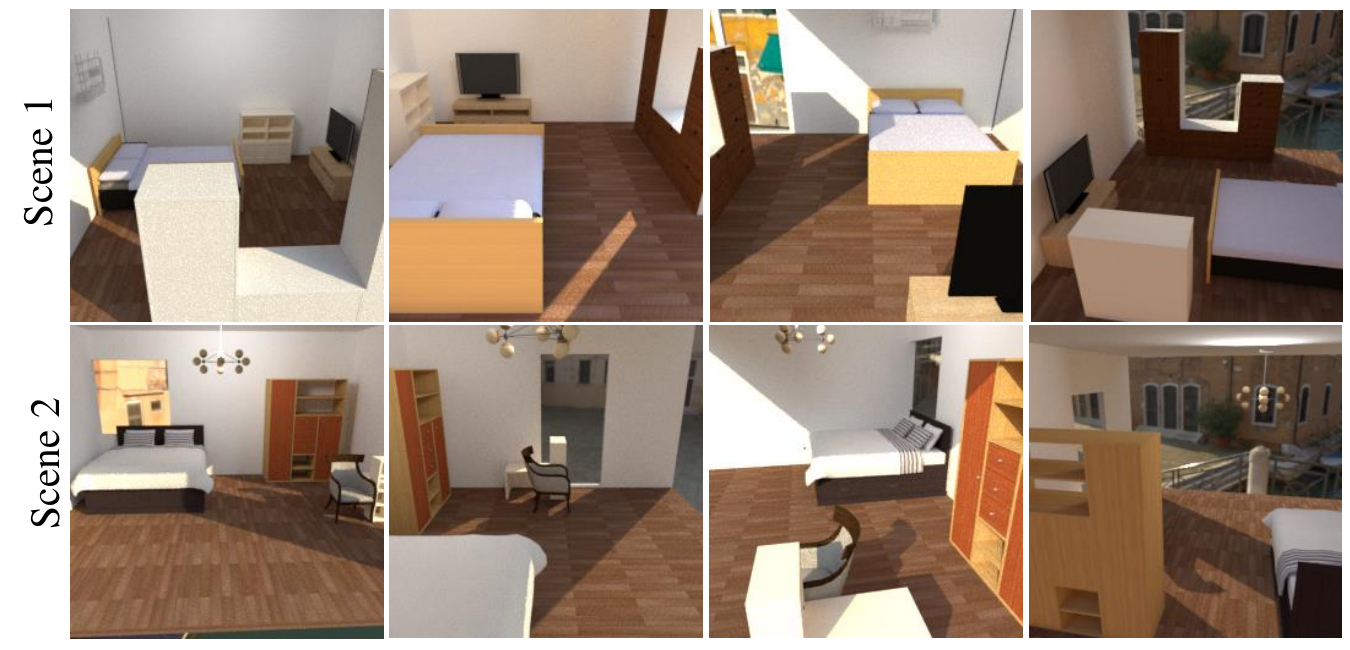}
  \vspace{-15pt}
  \caption{Multiple views of synthesized scenes.}
  \vspace{-10pt}
  \label{multiview}
\end{figure}

\myparagraph{Results.} 
The results can be seen in Table~\ref{finetune_number}. We also perform qualitative evaluations on layouts in Figure~\ref{finetunevisual}. As expected, the initial proposed layout shares the same scene graph with the target layout, the action location of the objects can be different from the target, because our layout synthesis is conditioned on scene graph only. After optimization, we are able to fit the target layout reasonably well. Multiple views of synthesized scenes are shown in Figure~\ref{multiview}.

\myparagraph{Analyzing the latent space.}

\begin{figure}[t!]
  \includegraphics[width=\linewidth]{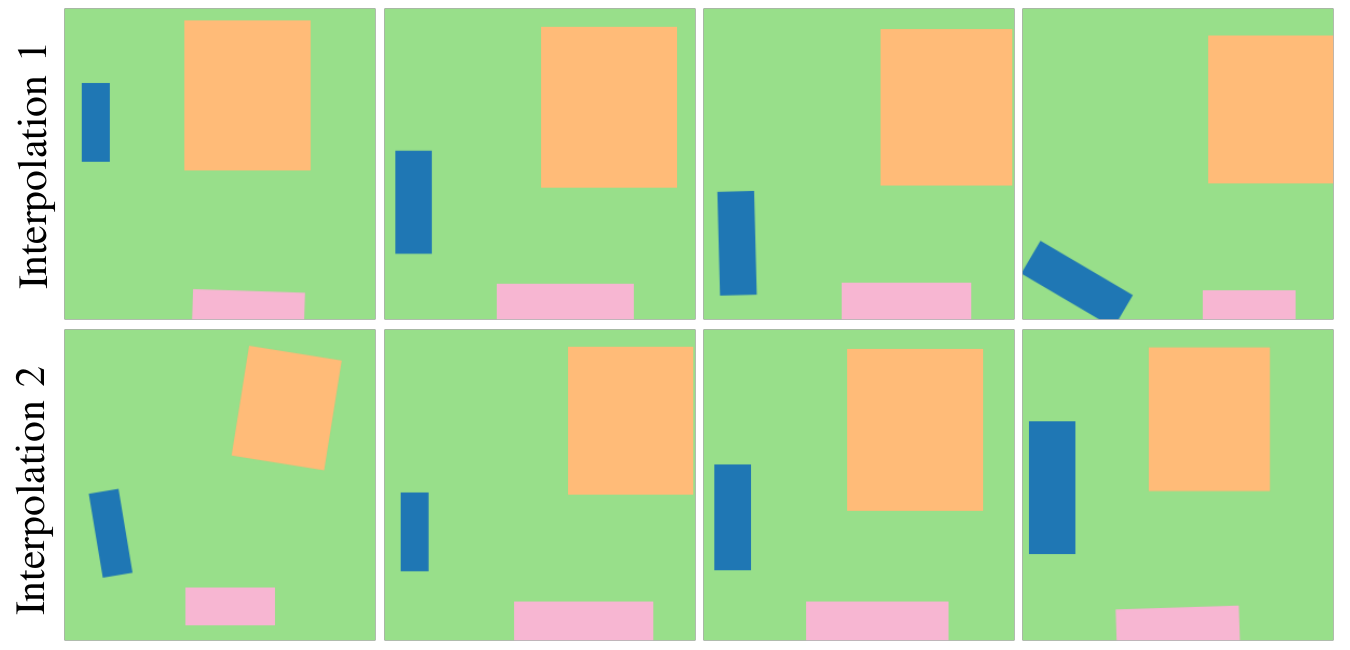}
  \vspace{-2em}
  \caption{Top down visualization of a room as we linearly interpolate between the latent vector representing layouts for the same scene graph.}
  \vspace{-1em}
  \label{interp}
\end{figure}

\begin{figure*}[t]
    \centering
    \includegraphics[width=\textwidth]{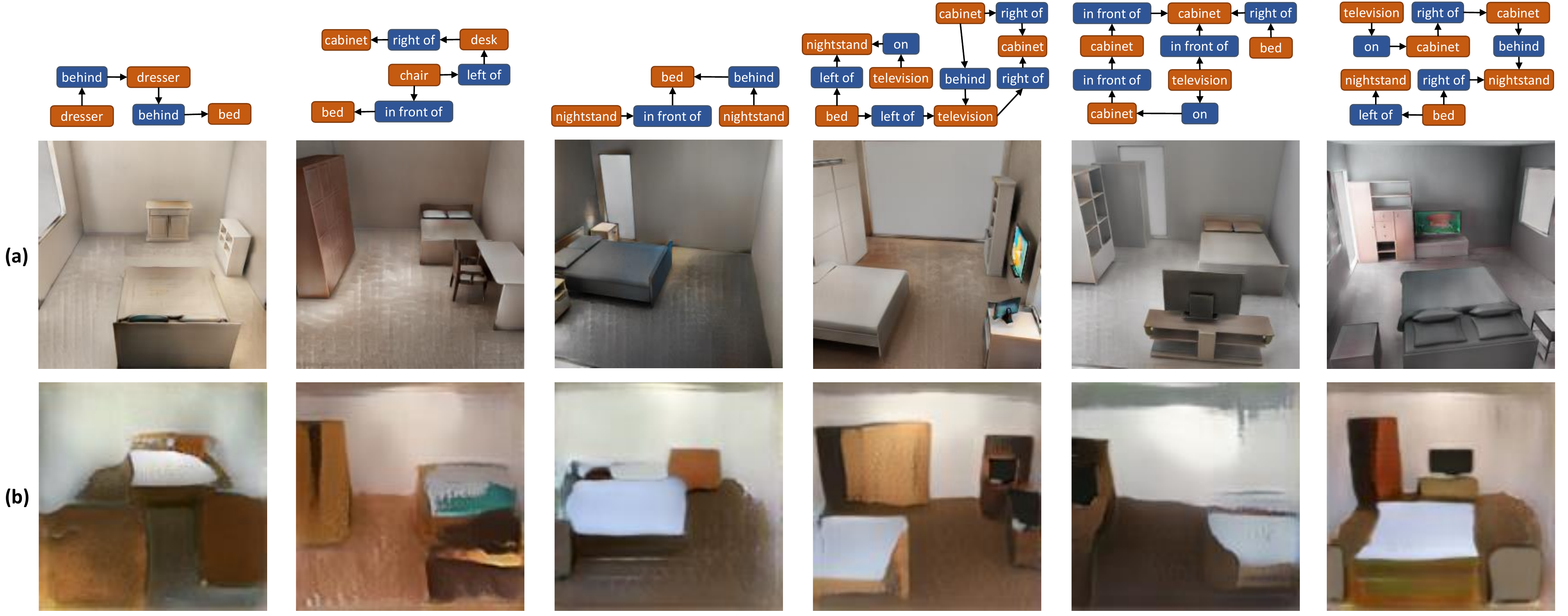}
    \vspace{-15pt}
    \caption{Qualitative results for conditional image synthesis. Top: input scene graph; Middle: images generated by our model; Bottom: images generated by Johnson~\etal\cite{johnson2018image}, which does not incorporate 3D information. Our model generates better results via its understanding of 3D scene layout.}
    \vspace{-15pt}
    \label{fig:main}
\end{figure*}
\begin{figure}[t]
  \includegraphics[width=\linewidth]{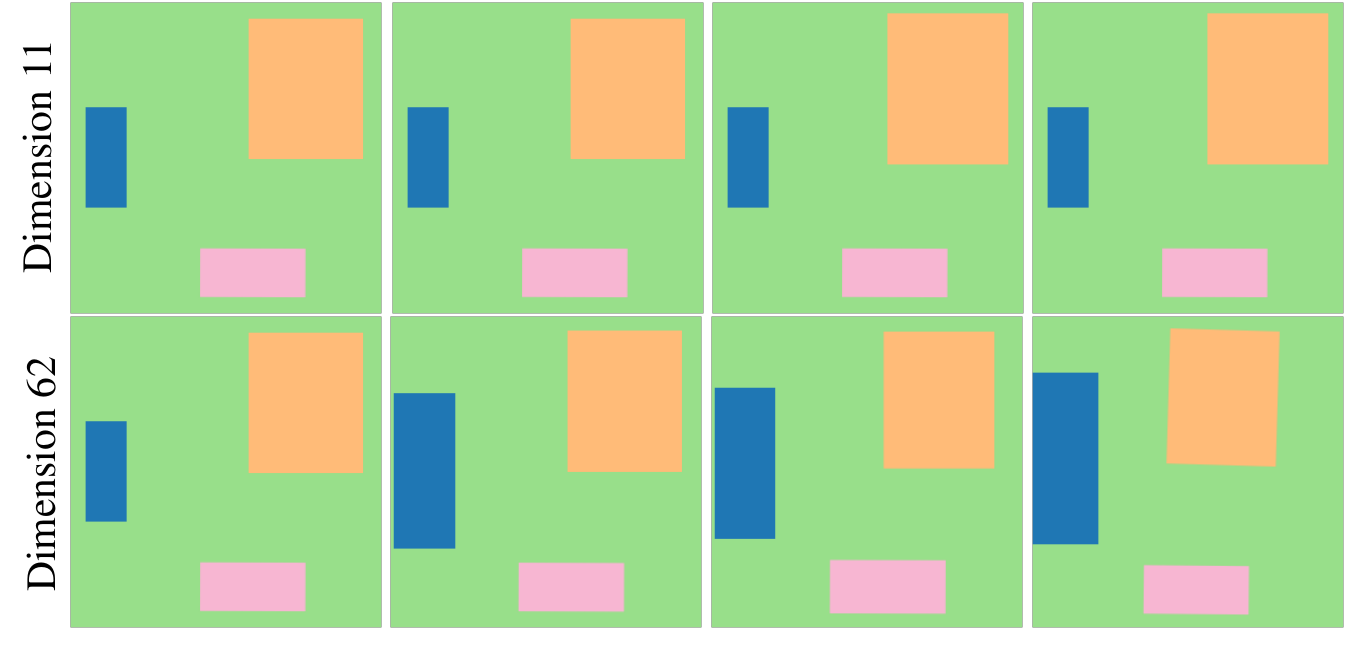}
  \vspace{-20pt}
  \caption{Top down visualization of a room as we manipulate individual dimensions of latent vector for bed object (orange). For dimension 11, note bed elongation.}
  \vspace{-10pt}
  \label{latent}
\end{figure}

We examine the latent representation of the scene layouts. In Figure~\ref{interp}, we demonstrate that the layout of objects can smoothly change as we interpolate two random latent vectors. In Figure~\ref{latent}, we demonstrate the effect of manipulating dimension $11$ and $62$ of the latent vector for the bed object.

\myparagraph{User study.} We randomly sample 300 scene graphs and, for each, generate five layouts using the GCN+noise model and five using our 3D-SLN model, respectively. We then present the layouts in a top down view along with the scene graph in sentences to subjects on Amazon Mechanical Turk, asking them which set of layouts is more diverse. Each subject is shown twelve scene graphs. 78.9\% of responses suggested layouts generated by 3D-SLN be more diverse. 

\myparagraph{Failure cases.}

\begin{figure}[t]
  \includegraphics[width=\linewidth]{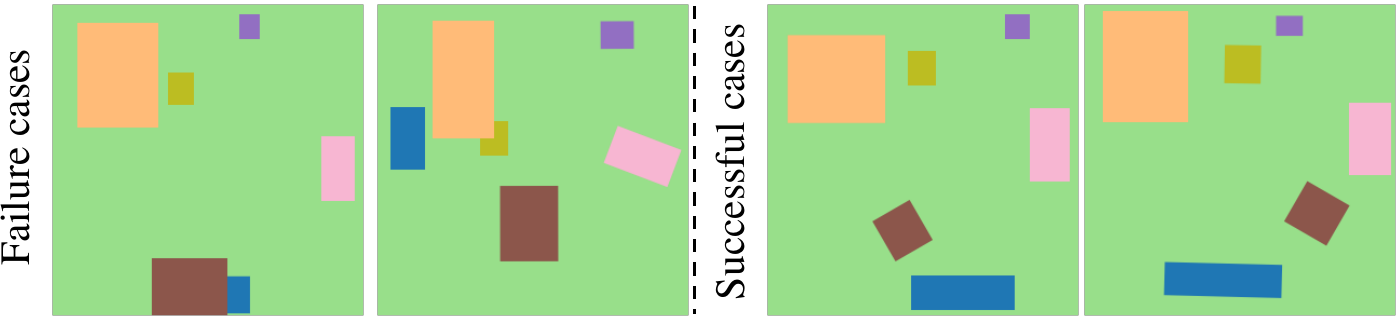}
  \vspace{-15pt}
  \caption{Top down visualization of failure cases (left of dotted line) compared to a good layout. All layouts are synthesized from the same scene graph.}
  \vspace{-10pt}
  \label{fail}
\end{figure} 

As shown in Figure~\ref{fail}, scene synthesis performance decreases when a graph contains too many objects that might overlap. As part of future work, this could be improved during training by adding an adversarial loss, or during inference by rejecting implausible layouts with the use of physical simulation as in~\cite{du2018learning}, or by performing  simple collision detection on generated layouts. 

\section{Applications}

Our scene graph--based layout synthesis algorithm enables many downstream applications. In this section, we show results on scene graph--based image synthesis, sentence-based scene layout synthesis, and exemplar-based scene layout synthesis.

\subsection{Scene Graph--Based Image Synthesis}

As our model produces not only 3D scene layouts, but also 2.5D sketches and semantic maps, we train an image translation network, SPADE~\cite{park2019semantic}, that takes in depth and semantic maps and synthesizes an RGB image. 
Training data for the SPADE model, including RGB, depth, and semantic maps, are all taken from the Structured3D dataset~\cite{zheng2019structured3d}, and are randomly cropped to 256$\times$256 after we resize the longest edge to 480 pixels. The training dataset consists of 82,838 images total. We compare our model with the state-of-the-art, scene graph-to-image model~\cite{johnson2018image}. 

Results are shown in Figure~\ref{fig:main}. The images generated by our model are sharp and photo-realistic, with complex lighting. Meanwhile, the baseline~\cite{johnson2018image} can only generate blurry images, where sometimes the objects are hardly recognizable and fail to preserve the 3D structure.

\begin{figure*}[t!]
    \centering
    \includegraphics[width=0.85\linewidth]{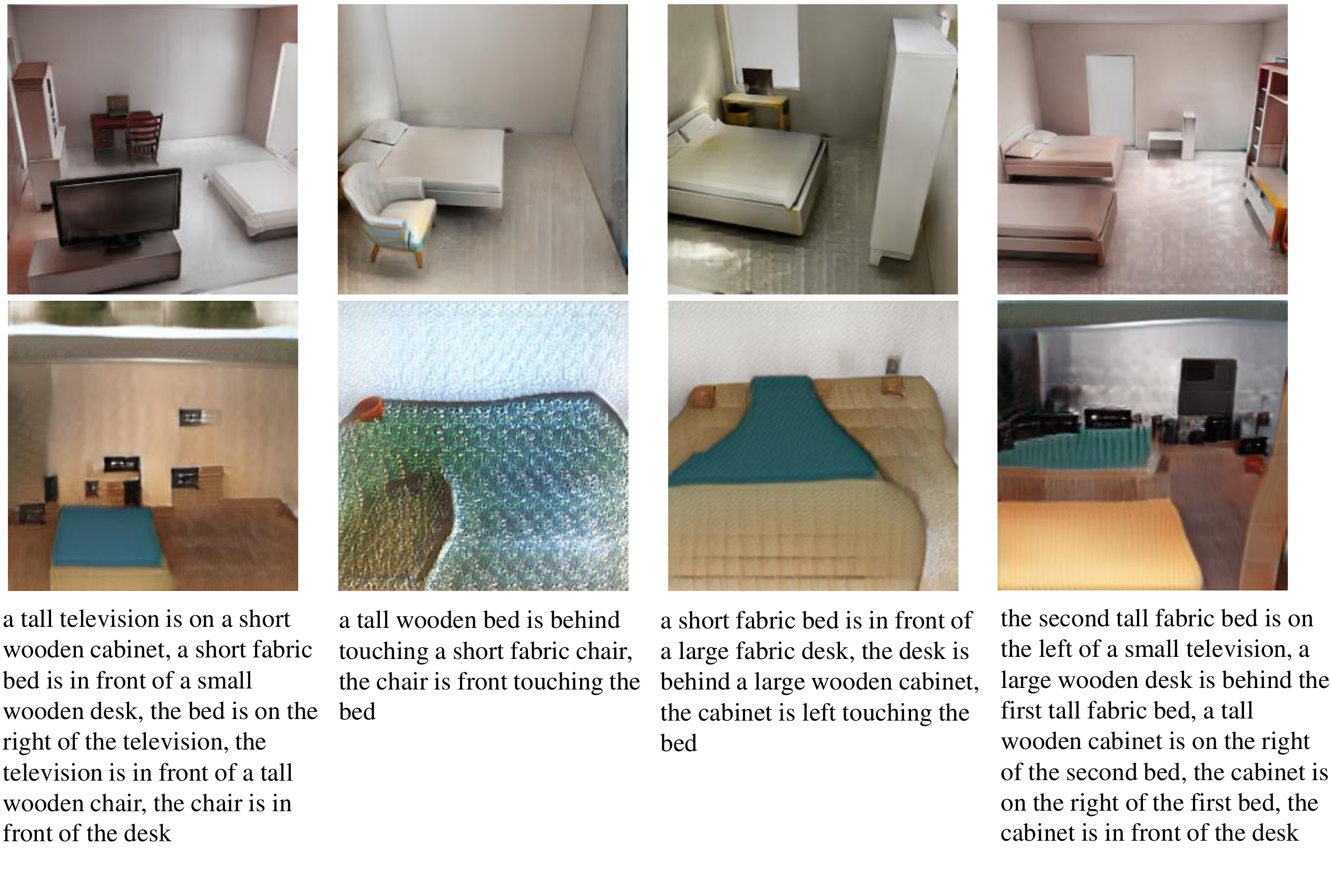}
    \vspace{-2em}
    \caption{Comparison with AttnGAN~\cite{xu2018attngan}. The images on the top row are generated by our model, while the images on the bottom row are generated by AttnGAN~\cite{xu2018attngan}. In comparison, our model generates higher-quality scenes.}
    \vspace{-1.5em}
    \label{attngan}
\end{figure*}

\subsection{Sentence-Based Scene Layout Synthesis} 
 
Conventional text-to-image synthesis methods use a text encoder to convert an input sentence into a latent code, which is then fed into a conditional GAN to generate an image. However existing methods only work when the input sentence has only one or a few objects. The task becomes more challenging when input text consists of multiple objects and contains complex relationships. We compare our approach against AttnGAN~\cite{xu2018attngan}, the state-of-the-art image synthesis algorithm that takes in sentences as input.

Qualitative results are shown in Figure~\ref{attngan}. As AttnGAN suffers from deterioration when there are too many objects, we have constricted each individual description to have at most five sentences. Our 3D-SLN generates more realistic images compared with AttnGAN.

\subsection{Exemplar-Based Scene Layout Synthesis}

Our model can also be used to reconstruct and create new layouts based on an example image. We use \textit{Cooperative Scene Parsing}~\cite{huang2018cooperative} to predict object classes and 3D bounding boxes from an image. For our purposes, we test on bedroom images sampled from the SUN RGB-D dataset~\cite{song2015sun}. After extracting 3D bounding boxes for each object, we infer a 3D scene graph with the same object classes and relationships that our model is trained on. This scene graph is sent to our model to generate layouts that observe the relational constraints present in the scene graph. 

\begin{figure}[t!]
  \centering
  \vspace{5pt}
  \includegraphics[width=\linewidth]{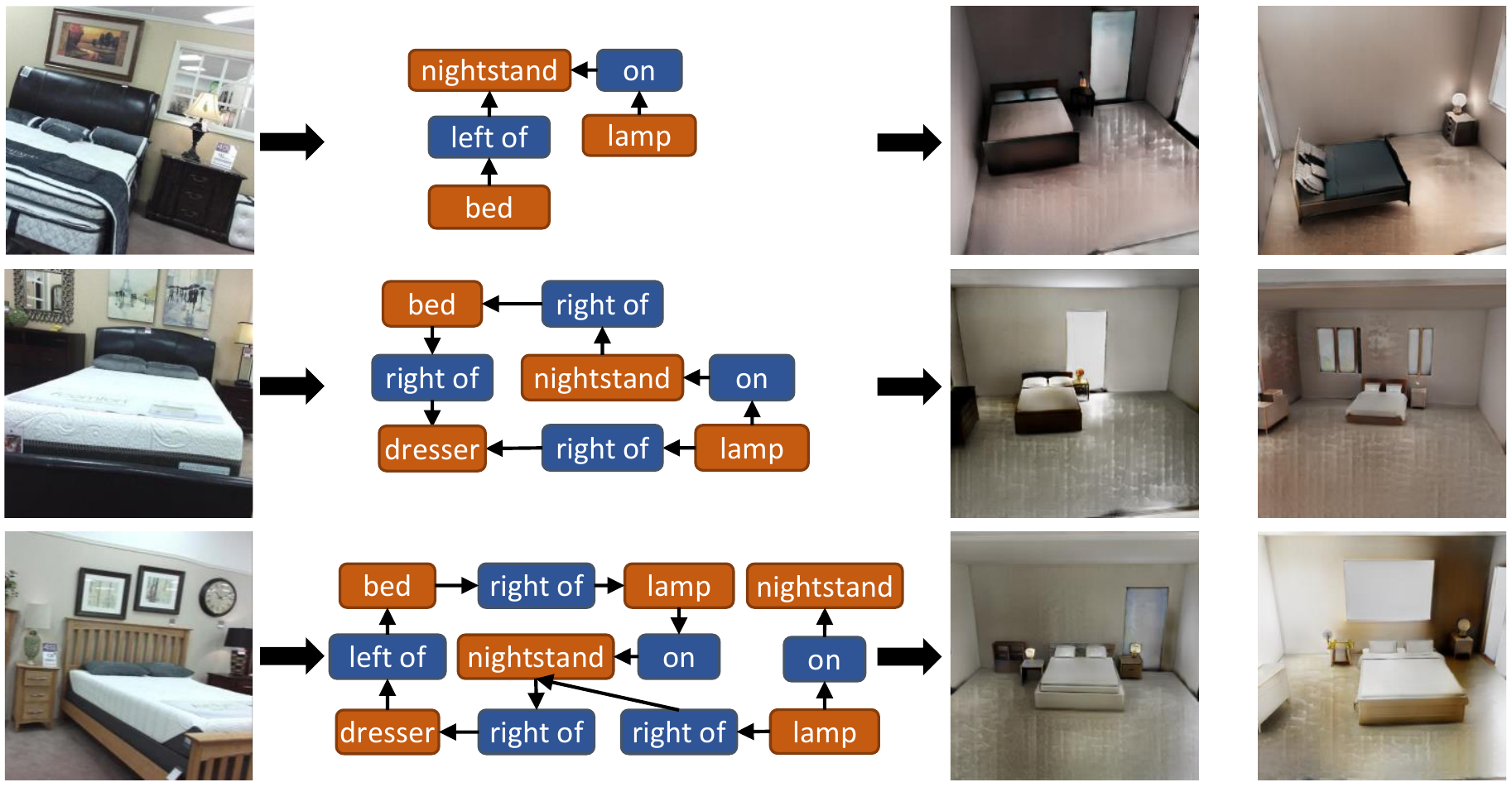}
  \vspace{-15pt}
  \caption{On the left most column are images of the class `bedroom' from the SUN RGB-D dataset~\cite{song2015sun}. 3D bounding boxes are calculated per object, and are fed to a rule-based parser, which generates the relationships and creates a scene graph. The scene graph is then fed to our 3D-SLN to generate diverse layouts. Final images are rendered with SPADE~\cite{park2019semantic}.}
  \label{fig:exemplar}
  \vspace{-15pt}
\end{figure}

We present some qualitative results in Figure~\ref{fig:exemplar}. Our model is not only capable of recovering the original layout in the example image, but it can also create new layouts according to the scene graph (notice the different locations and rotations of the bed and nightstand).

\section{Conclusion}

In this paper, we have introduced a novel, stochastic scene layout synthesis algorithm conditioned on scene graphs. Using scene graphs as input allows flexible and controllable scene generation. Experiments demonstrate that our model generates more accurate and diverse 3D scene layouts compared with baselines. Our model can also be integrated with a differentiable renderer to refine 3D layout conditioned on a single example. Our model finds wide applications in downstream scene layout and image synthesis tasks. We hope our work will inspire future work in conditional scene generation.

\vspace{6pt}
\noindent {\bf Acknowledgements.}
This work is supported by NSF \#1447476, ONR MURI N00014-16-1-2007, and NIH T90-DA022762.

{\small
\bibliographystyle{ieee_fullname}
\bibliography{egbib}
}
\newpage
\setcounter{section}{0}
\setcounter{table}{0}
\setcounter{figure}{0}
\onecolumn
\begin{center}\textbf{\Large{Supplementary material}}\end{center}
\section{Relational definitions}
We define the relationships in the 3D scene graph as listed in Table \ref{tbl:relationships}. Assume that $X$ and $Y$ define the ground plane, and $Z$ is the vertical axis.

\begin{table*}[h!]
	\begin{center}
	\begin{tabular}{p{3cm}p{3.5cm}p{10cm}}%
		\toprule
		\multirow{27}{*}{Relationships}
    & {On} & An object $o_i$ is \textit{\textbf{on}} $o_j$ when \newline \texttt{rel}$(o_i, o_j) = \text{inside}$ and \newline $\lvert{(\texttt{center}_{Z_j}-\texttt{center}_{Z_i}), \frac{\texttt{max}_{Z_j}-\texttt{min}_{Z_j}+\texttt{max}_{Z_i}-\texttt{min}_{Z_i}}{2} } \rvert \leq 0.05$ \\
		 \cmidrule{3-3}
    & {Left of} & An object $o_i$ is \textit{\textbf{left of}} $o_j$ when \newline $(\frac{3\pi}{4} \leq \theta_{i,j} \text{ Or } \theta_{i,j} \leq \ -\frac{3 \pi}{4})$ and $\texttt{IOU}(o_i, o_j) = 0$
    \\ 
    \cmidrule{3-3}
		& {Right of} & An object $o_i$ is \textit{\textbf{right of}} $o_j$ when \newline $-\frac{\pi}{4} \leq \theta_{i,j} \leq \frac{\pi}{4}$ and $\texttt{IOU}(o_i, o_j) = 0$\\ 
	\cmidrule{3-3}
		& {Behind} & An object $o_i$ is \textit{\textbf{behind}} $o_j$ when \newline $\frac{\pi}{4} \leq \theta_{i,j} \leq \frac{3\pi}{4}$ and $\texttt{IOU}(o_i, o_j) = 0$\\ 
	\cmidrule{3-3}
		& {In front of} & An object $o_i$ is \textit{\textbf{in front of}} $o_j$ when \newline $-\frac{3\pi}{4} \leq \theta_{i,j} \leq -\frac{\pi}{4}$ and $\texttt{IOU}(o_i, o_j) = 0$ \\
	\cmidrule{3-3}
		& {Right touching} & An object $o_i$ has its \textit{\textbf{right touching}} $o_j$ when \newline $(\frac{3\pi}{4} \leq \theta_{i,j} \text{ Or } \theta_{i,j} \leq \ -\frac{3 \pi}{4})$ and $0 < \texttt{IOU}(o_i, o_j)$ \\
	\cmidrule{3-3}
    & {Left touching} & An object $o_i$ has its \textit{\textbf{left touching}} $o_j$ when \newline $-\frac{\pi}{4} \leq \theta_{i,j} \leq \frac{\pi}{4}$ and $0 < \texttt{IOU}(o_i, o_j)$ \\
    \cmidrule{3-3}
    & {Front touching} &  An object $o_i$ has its \textit{\textbf{front touching}} $o_j$ when \newline $\frac{\pi}{4} \leq \theta_{i,j} \leq \frac{3\pi}{4}$ and $0 < \texttt{IOU}(o_i, o_j)$ \\
    \cmidrule{3-3}
    & {Behind touching} & An object $o_i$ has its \textit{\textbf{behind touching}} $o_j$ when \newline$-\frac{3\pi}{4} \leq \theta_{i,j} \leq -\frac{\pi}{4}$ and $0 < \texttt{IOU}(o_i, o_j)$ \\
    \cmidrule{3-3}
    & {Inside} & An object $o_i$ is \textit{\textbf{inside}} $o_j$ when \newline$\texttt{min}_{X_j} < \texttt{min}_{X_i}; \texttt{max}_{X_i} < \texttt{max}_{X_j}$\newline $\texttt{min}_{Y_j} < \texttt{min}_{Y_i}; \texttt{max}_{Y_i} < \texttt{max}_{Y_j}$ \\
    \cmidrule{3-3}
    & {Surrounding} & An object $o_i$ is \textit{\textbf{surrounding}} $o_j$ when \newline$\texttt{min}_{X_i} < \texttt{min}_{X_j}; \texttt{max}_{X_j} < \texttt{max}_{X_i}$\newline $\texttt{min}_{Y_i} < \texttt{min}_{Y_j}; \texttt{max}_{Y_j} < \texttt{max}_{Y_i}$\\

	\bottomrule
	\end{tabular}
	\end{center}
\caption{The list of relationships of our model, the existence of the \textit{\textbf{on}} relationship between any pair of objects prevents the occurrence of other relationships for a given pair.  Here $\theta_{i,j} =\atantwo{(Y_{i}-Y_{j},X_{i}-X_{j})}$.}
	\label{tbl:relationships}
\end{table*}

\newpage
\section{Attribute definitions}
We define the attribute of each object in the 3D scene graph as listed in Table \ref{tbl:attributes}.\\ \\

\begin{table}[H]
  \vspace{-0.2in}
	\begin{center}
	\begin{tabular}{p{3cm}p{3.5cm}p{10cm}}%
		\toprule
		\multirow{10}{*}{Attributes}
    & None & We assign an object no attributes, this the default
    \\
		 \cmidrule{3-3}
    & Tall & An object $o_i$ is \textit{\textbf{tall}} when $H_i$ is in the top $30\%$ of heights for objects of the same class
    \\ 
    \cmidrule{3-3}
		& Short & An object $o_i$ is \textit{\textbf{short}} when $H_i$ is in the bottom $70\%$ of heights for objects of the same class\\ 
	\cmidrule{3-3}
		& Large & An object $o_i$ is \textit{\textbf{large}} when $V_i$ is in the top $30\%$ of volumes for objects of the same class\\ 
	\cmidrule{3-3}
		& Small & An object $o_i$ is \textit{\textbf{small}} when $V_i$ is in the bottom $70\%$ of volumes for objects of the same class \\
	\bottomrule
	\end{tabular}
	\end{center}
	
	\caption{The list of attributes that can be assigned to an object. We denote the volume of $o_i$ as $V_i = (\texttt{max}_{X_i} - \texttt{min}_{X_i}) \times (\texttt{max}_{Y_i} - \texttt{min}_{Y_i}) \times (\texttt{max}_{Z_i} - \texttt{min}_{Z_i})$. We denote its height as $H_i = (\texttt{max}_{Z_i} - \texttt{min}_{Z_i})$.}
	\label{tbl:attributes}
\end{table}

\newpage
\section{Differentiable fine tuning: perspective and top down views}
We provide more visualizations for scene fine tuning using a differentiable renderer, presented in \textbf{section 4.4} of the main text. Here we visualize the top down views of the scene before fine tuning, the target scene for fine tuning, and the top down view of the scene after fine tuning.

\begin{figure*}[h!]
    \centering
    \includegraphics[width=0.88\linewidth]{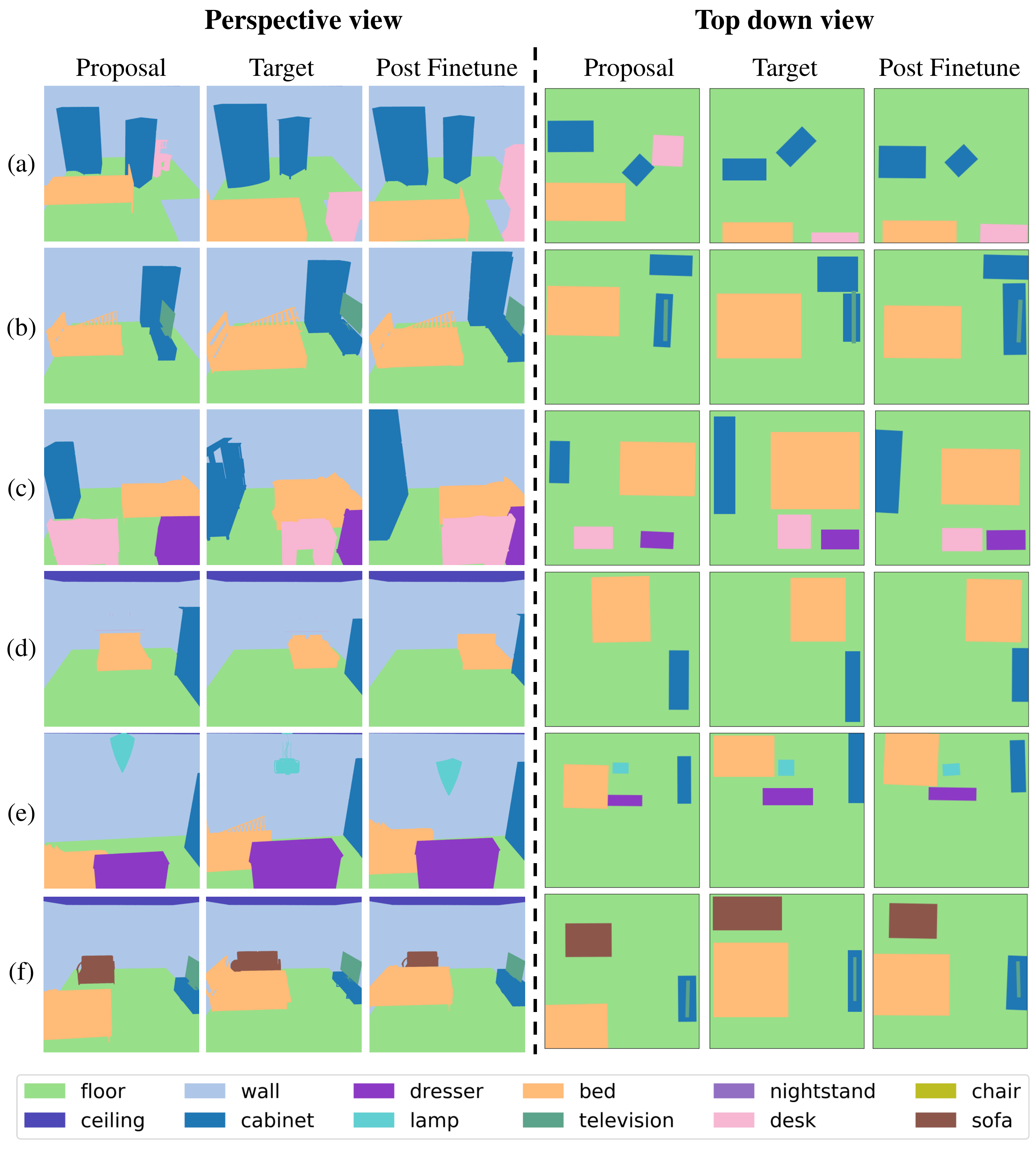}
    \caption{Object placements before and after differentiable fine tuning. The first three columns represent the semantic map from a camera placed inside the room, the latter three columns represent the top down view of the semantic map.  \textbf{(a)} Note the position of the cabinets and the desk; \textbf{(b)} Both cabinets are shifted closer to the corner, the bed is brought forward; \textbf{(c)} The desk and dresser are brought closer together, the cabinet is increased in size; \textbf{(d)} The bed is shifted right; \textbf{(e)} The bed is shifted to the back, the lamp is brought down; \textbf{(f)} The sofa is pushed backwards, and the bed is pushed backwards as well.}
    \label{fig:finetune}
\end{figure*}
\newpage
\section{Object distributions heatmaps for different scene graphs, top down views}
We provide a heatmap visualization demonstrating the diversity of our generated layouts conditioned on the input 3D scene graph. 

\begin{figure*}[h!]
    \centering
    \includegraphics[width=0.87\linewidth]{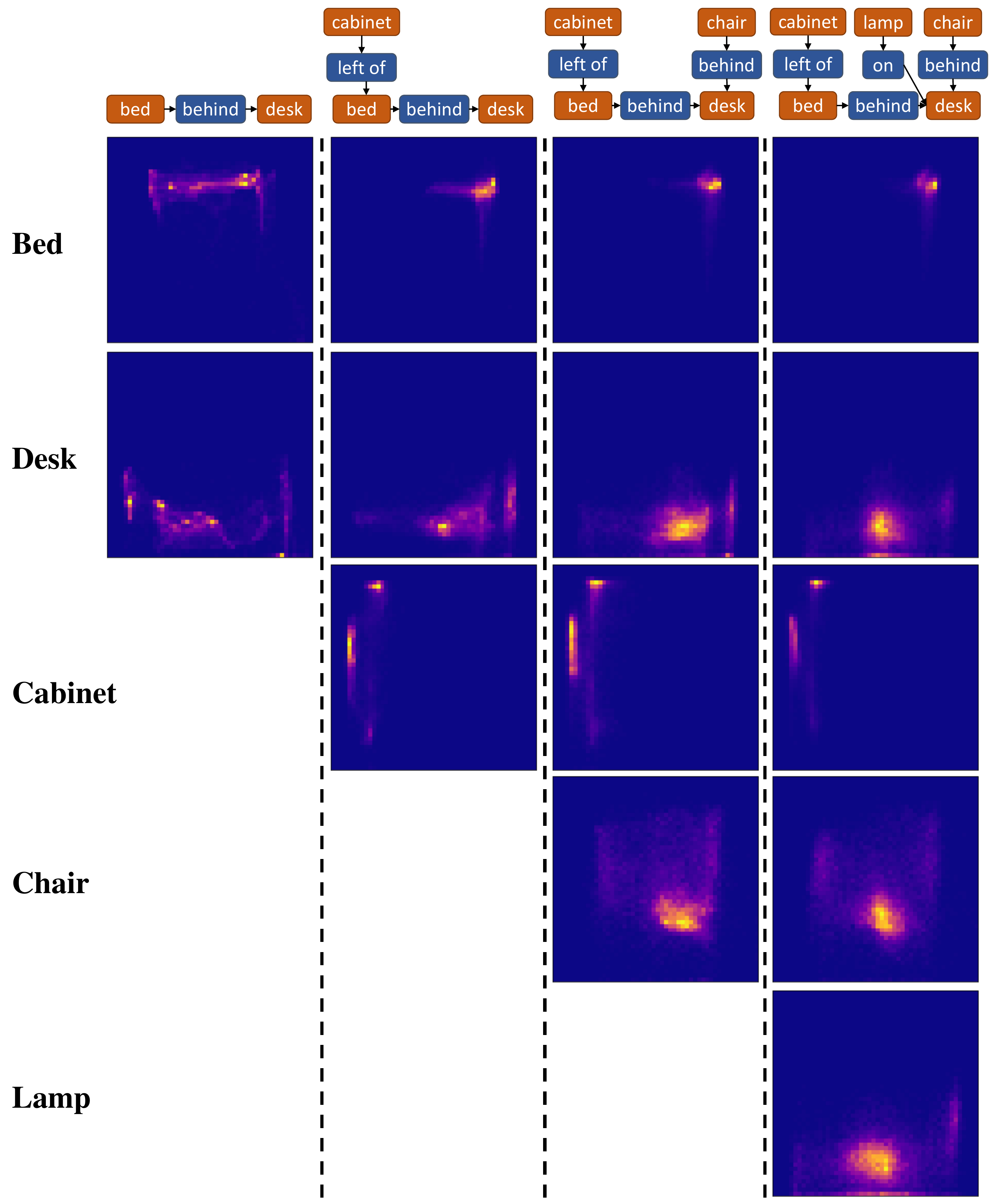}
    \caption{We visualize object centroids for different scene graphs. For each scene graph, we sample 20,000 latent vectors using the $\mu, \sigma$ computed from the training distribution. The density is visualized from a top down view. A scene graph with more objects produces more concentrated object placements.}
    \label{fig:heatmap}
\end{figure*}

\section{Example scenes from scene graph: perspective and top down views}
We provide the corresponding top-down view of the scenes for the scene-graph based image synthesis experiment, described in \textbf{Section 5.1} of the main text.

\begin{figure*}[h!]
    \centering
    \vspace{-0.1in}
    \includegraphics[width=0.66\linewidth]{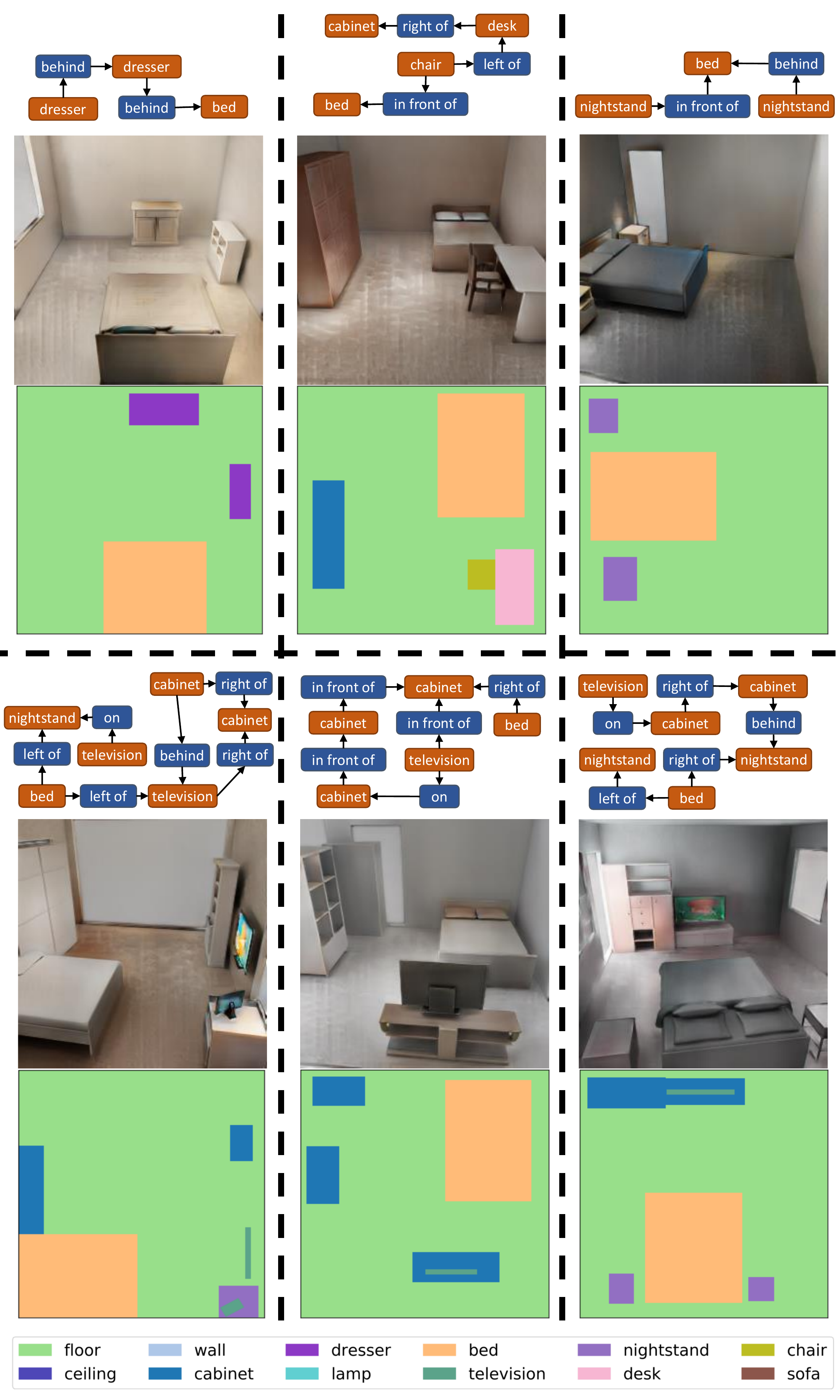}
    \caption{Perspective and top down views of synthesized scenes. For each block, the top images corresponds to the first row of Figure 7 in the main text, the second image represents the top down view of the scene. The view in perspective image is captured from a camera placed at the bottom edge of top down visualization, looking towards the center of the room.}
    \label{fig:finetune}
\end{figure*}

\end{document}